%% file: EGauthorGuidelines-conf-fin-cc-by-nc.tex
\documentclass{egpubl}
\usepackage{eg2023}

\ConferencePaper        %

\CGFccbyncnd

\usepackage[T1]{fontenc}
\usepackage{dfadobe}  

\usepackage{cite}  %
\BibtexOrBiblatex
\electronicVersion
\PrintedOrElectronic
\ifpdf \usepackage[pdftex]{graphicx} \pdfcompresslevel=9
\else \usepackage[dvips]{graphicx} \fi

\usepackage{egweblnk}
\usepackage{booktabs}
\usepackage{multirow}
\usepackage[]{graphicx}
\usepackage[]{xcolor}
\usepackage{amsmath}
\usepackage{float} 
\usepackage{amsfonts}
\usepackage{enumitem}
\usepackage{caption}
\usepackage{subcaption}

\definecolor{amethyst}{rgb}{0.6, 0.4, 0.8}
\definecolor{darkpastelgreen}{rgb}{0.01, 0.75, 0.24}
\definecolor{amber}{rgb}{1.0, 0.75, 0.0}
\definecolor{cadmiumorange}{rgb}{0.93, 0.53, 0.18}
\definecolor{lawngreen}{rgb}{0.49, 0.99, 0.0}
\definecolor{limegreen}{rgb}{0.2, 0.8, 0.2}
\definecolor{neongreen}{rgb}{0.22, 0.88, 0.08}
\definecolor{amethyst}{rgb}{0.6, 0.4, 0.8}
\definecolor{darkpastelgreen}{rgb}{0.01, 0.75, 0.24}
\definecolor{greenbest}{RGB}{88,137,15}
\definecolor{redworst}{RGB}{137,15,27}

\newcommand{\REMOVE}[1]{{}}

\newcommand{\GreenColor}[1]{\textcolor{greenbest}{\textbf{#1}}}
\newcommand{\RedColor}[1]{\textcolor{redworst}{\textbf{#1}}}

\gdef\etal{\textit{et al.}}

\newcommand{\Density}{\rho}
\newcommand{\dscene}{I_{\text{sc}}}
\newcommand{\vect}[1]{\mathbf{#1}}                %
\usepackage{egweblnk} 

\title[How Will It Drape Like?  Capturing Fabric Mechanics from Depth Images]{How Will It Drape Like? \\Capturing Fabric Mechanics from Depth Images}
\author[C. Rodriguez-Pardo, M. Prieto-Martin, D. Casas, E. Garces]
{\parbox{\textwidth}{\centering Carlos Rodriguez-Pardo$^{1,2,3}$\orcid{0000-0001-6121-7738}, Melania Prieto-Martin$^{2}$\orcid{0000-0002-4053-3720}, Dan Casas$^{2}$\orcid{https://orcid.org/0000-0002-3664-089X}, Elena Garces$^{1,2}$\orcid{0000-0003-3509-8485}
	}
	\\
	{\parbox{\textwidth}{\centering $^1$SEDDI, Madrid, Spain\\
			$^2$Universidad Rey Juan Carlos, Madrid, Spain\\
			$^3$Universidad Carlos III de Madrid, Spain
		}
	}
}

\begin{document}
	
	\teaser{
		\vspace{-28pt}
		\includegraphics[width=\linewidth]{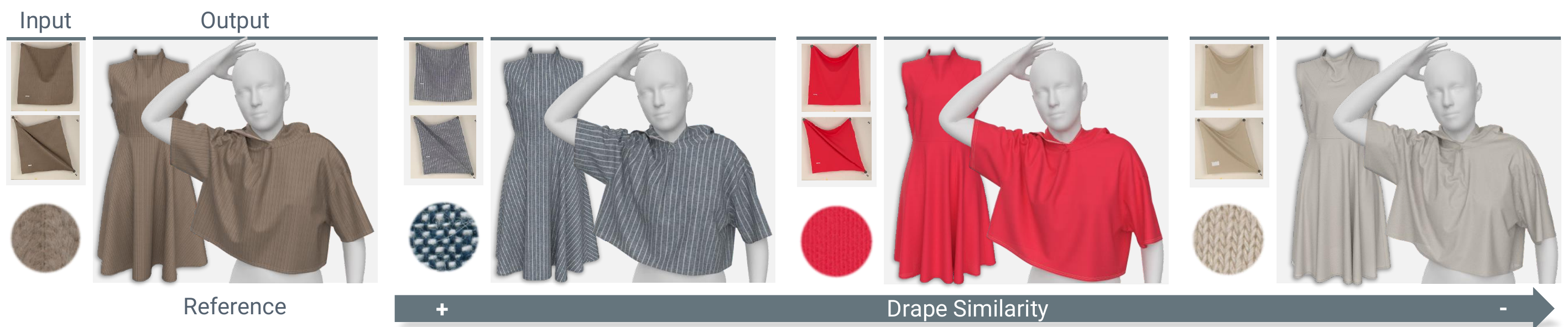}
		\caption{From just two depth images of a fabric sample casually placed in two specific configurations (input), we accurately infer the corresponding set of mechanical parameters of the material.
			Estimated parameters can be used in a cloth simulator, enabling to visualize the overall drape of any garment (output). 
			Our method further introduces a novel perceptually-validated drape similarity metric, which enables sorting materials based on their final drape.}
		\label{fig:teaser}
		\vspace{10pt}
	}
	
	\maketitle
 	
	\begin{abstract}
		We propose a method to estimate the mechanical parameters of fabrics using a casual capture setup with a depth camera.
		Our approach enables to create mechanically-correct digital representations of real-world textile materials, which is a fundamental step for many interactive design and engineering applications.
		As opposed to existing capture methods, which typically require expensive setups, video sequences, or manual intervention, 
		our solution can capture at scale, is agnostic to the optical appearance of the textile, and facilitates fabric arrangement by non-expert operators.
		To this end, we propose a sim-to-real strategy to train a learning-based framework that can take as input one or multiple images and outputs a full set of mechanical parameters.
		Thanks to carefully designed data augmentation and transfer learning protocols, 
		our solution generalizes to real images despite being trained only on synthetic data, hence successfully closing the sim-to-real loop. 
		Key in our work is to demonstrate that evaluating the regression accuracy based on the similarity at parameter space leads to an inaccurate distances that do not match the human perception.
		To overcome this, we propose a novel metric for fabric drape similarity that operates on the image domain instead on the parameter space, allowing us to evaluate our estimation within the context of a similarity rank. 
		We show that out metric correlates with human judgments about the perception of drape similarity, and that our model predictions produce perceptually accurate results compared to the ground truth parameters.

		

	\end{abstract}  
	
	\input{sections_cc/introduction}

	\input{sections_cc/relatedwork}

	\input{sections_cc/overview}

	\input{sections_cc/dataset2}

\input{sections_cc/method}

	\input{sections_cc/evaluation_params}

	\input{sections_cc/distancemetric}

	\input{sections_cc/results}

\input{sections_cc/resultsqualitative}

	\input{sections_cc/conclusions}

	\bibliographystyle{eg-alpha-doi}  
	\bibliography{references}

\end{document}

%% file: sections_cc/introduction.tex
\section{Introduction}\label{sec:introduction}

Creating accurate digital representations of real-world materials, or \textit{Digital Twins}, is crucial for enabling realistic 3D visualizations suitable for interactive design and predictive engineering. Some industries, like fashion or textile manufacturing, further require these methods to work at a scale to cope with the fast pace of the current production workflows. 
However, digitalizing cloth is challenging due to the high variability and type of fabric samples, where the fabric composition, the microstructure, or the finishing play crucial roles in the perceived appearance. 

While casual systems which obtain optical appearance have long been a focus of research, comparably less attention has been paid to estimating \textit{mechanical} properties. 
Indeed, capturing and simulating the mechanical behavior of cloth is challenging due to the complex interplay between the internal and external forces occurring in this type of physical system, which is highly sensitive to the environmental conditions. 
Nevertheless, with the current need to create virtual copies instantly, a casual setup able to produce automatic and accurate estimates --beyond having a set of \textit{presets} from which to manually choose the closest one-- of mechanical parameters could prove very valuable.

\begin{figure}[t]
	\centering

	\includegraphics[width=0.9\linewidth]{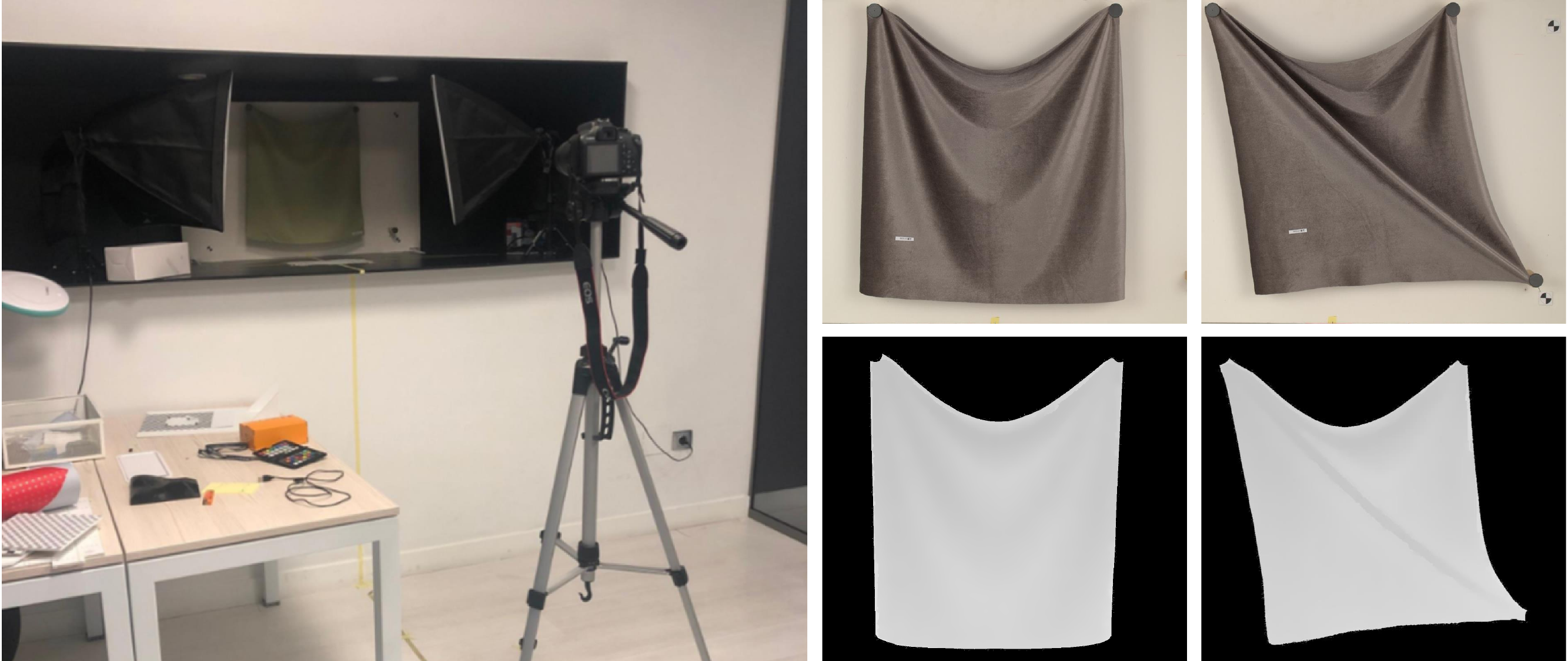}
 \vspace{-1mm}

	\caption{Capture setup, RGB (top), and depth (bottom) images for \textit{hanging} (left) and \textit{stretch} (right) scenes in rest position. Each scene conveys a different mechanical appearance of the fabric: \textit{hanging} exhibits the overall drape; \textit{stretch} exhibits an extra diagonal tension, which is key to understand the stretching properties.}
	\vspace{-5mm}
	\label{fig:simScenes}
\end{figure}

Previous methods are impractical for scalable and customizable workflows. 
Accurate fabric parameter acquisition systems require specialized and expensive devices~\cite{kawabata1980standardization, minazio1995fast, clapp1990indirect}, which are often slow and need skilled operators.
Existing casual capture setups use input video sequences~\cite{Bhat2003,bouman2013estimating,yang2017learning} or, even if they take a single image, might require manual user input~\cite{ju2020estimating}.
In this paper, we present a casual capture system that only requires taking two depth images of the textile posed in a static drape. Our capture setup does not require complex calibration, can be easily manipulated by non-expert operators, and is agnostic to the optical properties of the fabric thanks to leveraging depth images instead of RGB data.
Figure~\ref{fig:simScenes} illustrates our capture setup. It involves capturing the fabric with a depth camera in two relaxed positions: the \emph{hanging} scene, that conveys the drape when no force other than gravity is applied to it, and the \emph{stretch} scene, which provides cues on the stretching properties of the fabric.

We propose a learning-based method to instantly return the mechanical parameters given static depth images and the fabric density as input. 
Our method relies on a sim-to-real strategy~\cite{zhu2021survey}, leveraging transfer learning and building on a dataset of physical fabrics digitalized with precision equipment. 
Our model is trained solely with synthetic data, and thanks to carefully designed policies of data augmentation and neural features, it can generalize to real-world scenarios. 
Under the hood, our approach leverages a custom architecture that enables a flexible design, that can take one or multiple images as input, enhancing its performance when more data is available. 
We perform an extensive evaluation by means of ablation studies and by measuring aggregated neural network saliency maps, which show that some scenes are more informative than others for predicting the mechanical properties of the fabric. 
Furthermore, we demonstrate the performance of our model on real-world captured samples, showcasing our system's generalization capabilities. 

Key to our work is to demonstrate that evaluating the prediction accuracy of mechanical properties using typical error metrics, such as the Mean Absolute Error (MAE) on the parameter space, leads to inaccurate distances that do not match the human perception. 
We identify that such mismatches occur due to two factors: first, the parameter space is not bijective --i.e., different set of parameters might convey the same drape--; and second, a numerical error in a parameter does not necessarily correlate with what we perceive as an error. 
To address this shortcoming, common in all existing works, and inspired by previous work on similarity metrics for material appearance~\cite{lagunas2019similarity}, natural images~\cite{zhang2018unreasonable} or illustration~\cite{garces2014similarity}, we propose an image-based metric that measures differences on the mechanical behavior of textiles taking into account the overall drape.
We validate that our metric agrees with human perception,
and it and can be used to sort materials by drape similarity with respect to a reference fabric.

Using our similarity metric, we finally validate that the estimations of our method correlate with human judgments about drape similarity, and that our model predictions produce perceptually accurate results compared to the ground truth parameters. 
All in all, our approach makes an important step towards solving sim-to-real problems for mechanical estimation since it shows that simulated cloth using inferred parameters maximize the similarity with respect to real-world target fabrics.

%% file: sections_cc/relatedwork.tex
\section{Related Work}\label{sec:relatedwork}

\begin{figure*}[h!]
	\centering
	\includegraphics[width=1\linewidth]{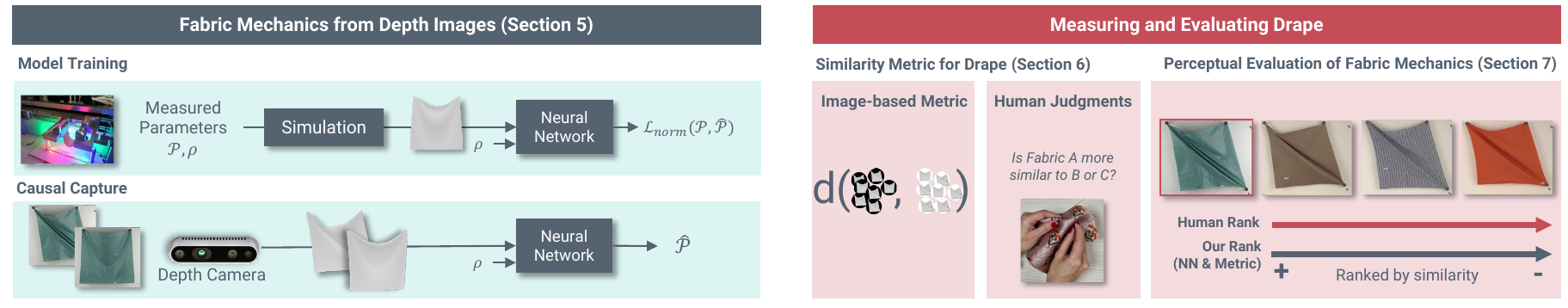}
	\vspace{-7mm}
	\caption{An overview of the main components of our method. We propose a technique to estimate fabric mechanics using depth images of \textit{hanging} and \textit{stretch} scenes as input. To validate the error of our estimations in a perceptual manner --accounting for the global drape--, we propose an image-based drape similarity metric which we validate with human judgments and can be used to sort fabrics by similarity. We show through several metrics that the estimations provided by our method using our similarity metric agree with those given by humans.}
 \vspace{-4mm}
	\label{fig:overview}
\end{figure*}

\subsection{Parameter Estimation Methods}

Estimating the mechanical properties of real fabric samples is a highly challenging problem for several reasons: 
the number of uncontrollable extrinsic factors (e.g., wind forces, initial state, collisions,  etc.) which affect the predictiveness of the physical simulation; 
the lack of a standard deformation model and parameter spaces; 
and the use of computation-intensive simulation methods. Accordingly, a wide range of strategies exist, aimed at overcoming these challenges.

\paragraph*{Measurement Devices.} It is common to combine optimization techniques with the output of testing devices to find 
the optimal set of parameters which best explain the observations \cite{magnenat2007measured,syllebranque2008estimation,volino2009simple,wang2011data,miguel2012data,clyde2017modeling}.
Existing technologies of this type
are diverse and, as discussed by Kuijpers \etal~\cite{kuijpers2020measurement}, lack of a clear standard. The Kawabata Evaluation System (KES) \cite{kawabata1980standardization} is perhaps one of the most well known, measuring 16 coefficients including bending, shearing, and tensile among others.
Despite its precision, this method was not widely adopted by the industry due to its lengthy processes and the need of expensive equipment. 
Consequently, several other methods tried to simplify and unify the methodology with partial success according to some studies~\cite{luible2008simulation,power2013fabric}: the Fabric Assurance by Simple Testing (FAST)~\cite{minazio1995fast}, the Fabric Touch Tester (FTT), the CLO Fabric Kit 2.0, the Fabric Analyser by Browzwear (FAB), the Optitex Mark 10, and the cantilever principle~\cite{clapp1990indirect}.

\paragraph*{Reconstruction-Optimization Methods.} Another set of techniques jointly tackles the reconstruction and parameter optimization problems. 
By taking as input data from arbitrary real simulations (e.g., the cloth deforming on an avatar~\cite{yang2018physics}), they iteratively reconstruct and simulate the scene which better explains the observation. 
Bhat~\etal~\shortcite{Bhat2003} takes as input a video sequence of the cloth and use simulated annealing to optimize the parameters by measuring its folds. Yang~\etal~\shortcite{yang2015materialcloning} use multi-view stereo reconstruction to initialize the 3D shape. Runia~\etal~\shortcite{runia2020cloth} also introduces simulation steps to explain observed phenomena of cloth in the wind. They rely on similarity metrics computed on deep latent spaces to supervise the optimization of the parameters.
These methods require simulation steps embedded into the fitting processes, making them computationally expensive due to the high dimensionality of the parameter spaces. Recent differentiable simulation techniques\cite{liang2019differentiable,jaques2019physics,hu2019difftaichi,murthy2020gradsim,li2022diffcloth} have proven to be efficient ways to reduce the fitting burden by enabling the computation of gradients with respect to these parameters within latent spaces of neural networks, taking into account dynamics, self-collisions, and contacts.

\paragraph*{Data-Driven and Regression Methods.} The third set of methods avoids reconstructing the original 3D scene by working on an estimated feature space and leveraging previously simulated data and machine learning techniques. Our approach falls in this category. 
Taking videos as input has been explored by Bouman~\etal~\shortcite{bouman2013estimating} to recover stiffness and area weight using a descriptor of the image based on PCA and optical flow, and later by Yang~\etal~\shortcite{yang2017learning}, who leverage neural networks to extract image features used for regression. 
Davis~\etal~\shortcite{davis2015visual} estimated the same simulation parameters by exploiting imperceptible vibrations in high-speed video recordings.
Bi~\etal~\shortcite{bi2018estimating} further evaluated that humans also need fabric motion to understand its stiffness.
Friction coefficients have been estimated using reflectance values~\cite{zhang2016friction} or dynamic videos of cloth sliding through a surface~\cite{rasheed2020learning}
Instead of regressing the parameters, Huber~\etal~\shortcite{huber2017cloth} find the most similar cloth in a database using motion descriptors.
A different approach only using a single image of the \emph{Cusick drape} was followed by Ju~\etal~\shortcite{ju2020estimating}, but it requires a 360$^{\circ}$ scan to reconstruct the target cloth, and a manually fitted Bezier curve to obtain the feature vector. In contrast, we just require a depth map that can be captured easily. Concurrent work~\cite{feng2022learning} uses multiple-view depth images as input to a trained regressor.

Our approach is inspired by these ideas; however, we do not require optimization --providing instant estimation of the parameters-- and leverage neural features to understand and model fabric behavior in a semi-controlled setup.
We demonstrate that our approach works with two images as input to predict bending and stretching coefficients without requiring a full video of the piece of fabric.

\subsection{Pre-Trained Models and Transfer Learning}

Deep learning models typically require vast amounts of data for generalizing to unseen examples. When this amount of data is not possible to acquire, \emph{tranfer learning} techniques helps by re-using model parameters trained on a related task~\cite{kornblith2019better, raghu2019transfusion, bommasani2021opportunities}. 
These techniques include:
\emph{Fine-tuning} the weights of a pre-trained classification model~\cite{weiss2016survey, rattani2017fine, chen2018fine, wang2018interactive,reddy2019transfer, guo2019spottune, kolesnikov2020big}. Pre-training an image descriptor model on contrastive or self-supervised learning tasks, and use the activations of its last layer as input to the downstream task (\emph{Linear Probing})~\cite{chen2020simple, radford2021learning, chen2021empirical, he2022masked, kumar2022fine}. For domain adaptation problems, it is common to \emph{adapt} the internal representations of pre-trained CNNs so as to efficiency~\cite{rebuffi2017learning, rebuffi2018efficient,rodriguez2019personalised, pham2020study, li2022cross}. Inspired by these approaches, we design a model that leverages fine-tuning of a pre-trained image CNN classifier as a feature extractor, capable of processing depth images, and extend it to account for additional input variables, and handling multiple images at the same time during test.

\subsection{Similarity Metrics}

\emph{Full-Reference Image Quality Assessment} (IQA) aims to provide a single score which measures the amount of distortion between two images. 
Traditionally, these metrics leveraged low-level image statistics. \emph{PSNR} is commonly used for measuring image degradation, but correlates poorly with human perception~\cite{zhang2018unreasonable}. More sophisticated alternatives have been developed, including \emph{SSIM}, \emph{mSSIM}~\cite{wang2004image}, and others~\cite{zhang2011fsim,nafchi2016mean,zhang2014vsi,zhang2012sr,reisenhofer2018haar}.
Algorithms based on latent spaces of CNNs~\cite{gatys2016image} have been extended to better approximate human perception, for example, by training on a large pool of human evaluations~\emph{LPIPS}~\cite{zhang2018unreasonable}, or by other means \cite{ding2020image,prashnani2018pieapp}.

Besides, similarity metrics that measure abstract or complex concepts like style have been proposed for 3D furniture~\cite{lun2015elements}, illustration~\cite{garces2014similarity}, icons~\cite{lagunas2019learning}, product design~\cite{lun2015elements}, or material appearance~\cite{lagunas2019similarity}.
Unlike ours, these metrics require to be trained with human ratings, thus incurring a considerable cost to collect such information via user studies. Instead, our metric does not require specific training, leveraging an off-the-shelf image-based metric. Despite this, we show that our metric correlates with human judgments on the perception of fabric drape similarity and that can be used to evaluate the overall drape.

%% file: sections_cc/overview.tex
\section{Overview}\label{sec:overview}
Figure~\ref{fig:overview} presents an overview of our work. 
First, in Section~\ref{sec:method}, we introduce our novel solution to infer fabric mechanics directly from depth images.
As input, our approach only requires static depth images in two specific configurations, shown in Figure~\ref{fig:simScenes}, as well as the fabric density which can be easily obtained with conventional equipment. 
In Section~\ref{sec:dataset} we describe the datasets of \textit{synthetic} and \textit{real} samples, with mechanical ground truth parameters, used train and evaluate our regressor.

The quantitative evaluation suggests that
our method is able to
estimate the mechanics within a certain error. 
However, since a direct interpretation of that error is not human-friendly, we propose a method to evaluate the overall drape in the context of a real scene. In Section~\ref{sec:distance_metric}, we introduce our image-based similarity metric for drape, which takes as input renders of the chosen scenes and provides a relative value that is useful to compare the drape of different fabrics. With a user study, explained in Section~\ref{sec:humanjudgements}, we validate that our metric agree with human preferences on the global perception of fabric drape. 
Also, in Section~\ref{sec:image-vs-human}, we evaluate our capture method using our drape similarity metric and compare it with human judgments. We effectively validate that our estimations agree with human assessments and provide several qualitative examples in Section~\ref{sec:qualitativeresults}.

%% file: sections_cc/dataset2.tex
\section{Datasets}\label{sec:dataset}

We develop two different datasets of depth images, which we use at different steps of the pipeline to train and evaluate our models: a \textit{synthetic} dataset, generated using physics-based cloth simulation;
and \textit{real} dataset, generated using images of real fabric samples.
In both datasets, a sample consists of a depth image of a fabric simulated in a scene for which we have the corresponding mechanical parameters, namely: bending~\cite{grinspun2003discrete} and stretch~\cite{volino2009simple} in the warp, weft, and bias directions and the fabric density, $\{$\emph{kStretchWarp}, \emph{kStretchWeft}, \emph{kStretchBias}, \emph{kBendinghWarp}, \emph{kBendingWeft}, \emph{kBendingBias}, $\Density \} \in \mathbb{R}^7$.
We support two different static configurations for the scenes:  
\textit{hanging}, which exhibits the overall drape; and \textit{stretch}, which exhibits diagonal tension and is key to understanding the stretching properties. Figure~\ref{fig:simScenes} depicts examples of each configuration.

\paragraph*{Simulated Dataset.} To train our model, we generate a synthetic dataset by simulating fabrics in a virtual scenario, replicating the \emph{hanging} and \emph{stretch} configurations.
We use a standard simulator, similar to ARCSim~\cite{narain2012adaptive}, with a quadratic strain, a linear strain/stress relationship, and standard definitions for bending and stretch \cite{grinspun2003discrete,volino2009simple}.
To model highly anisotropic fabrics, we use three parameters for warp, weft, and bias. Note that the model used for stretch already has some nonlinear behavior (quadratic strain), but more parameters (one per direction) are required to control the nonlinearity of the forces. Thickness is not included as it is implicitly accounted for in the other parameters. Contrarily, density is required as the simulator is dynamic. In a static one, it could be dropped after normalizing the other parameters. 

\begin{figure}[t]
	\centering
	\includegraphics[width=0.90\linewidth]{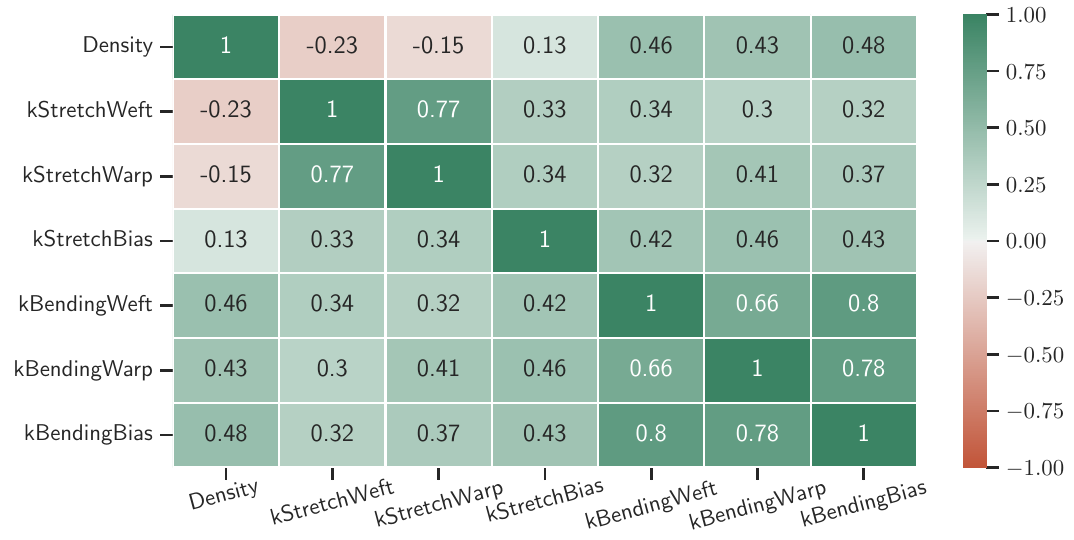}
 \vspace{-2mm}
	\caption{Spearman correlation matrix between parameters of our synthetic dataset.}
	\label{fig:dataset-statistics}
\end{figure}

\begin{figure}[t]
	 \vspace{-2mm}
	\centering
	\includegraphics[width=0.95\linewidth]{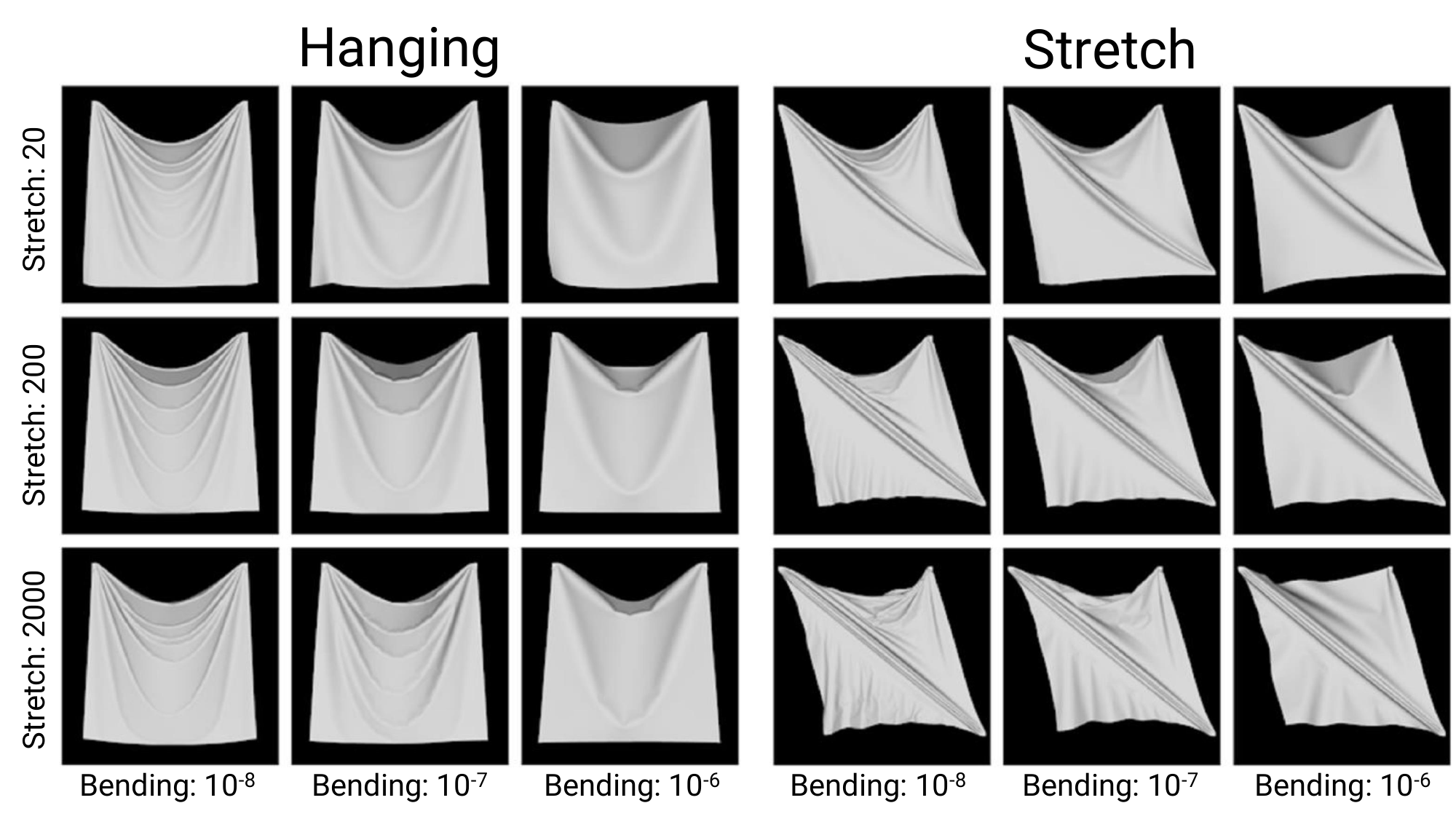}	
 \vspace{-2mm}
	\caption{Sweep of simulation parameters for hanging and stretch scenes. For \textit{kBending}: warp, weft, and bias have the same value, while for \textit{kStretch}, bias changes as 100, 144, 1000.}
 \vspace{-6mm}
	\label{fig:sweep}
\end{figure}

After simulation, we render the resulting mesh (discretized at 5mm/edge) with a white Lambertian material and extract the depth buffer.
To ensure that our synthetic dataset covers a wide range of materials, we densely sample the parameter space using a distribution of common mechanical parameters of real-world fabrics. 
Figure~\ref{fig:sweep} shows a sweep of parameters showcasing the variability of resulting drapes in the hanging and stretch scenes. 
To better understand potential relationships between the parameters, we compute the Spearman correlation $r$, shown in Figure~\ref{fig:dataset-statistics}, where we observe the higher correlation between the three kStretch coefficients and some correlation between the kBendingBias and the density.

\paragraph*{Real Dataset.} 

To evaluate our model, we test it with real data from images captured by the Intel RealSense SR300.
To this end, we casually hang $50\times50$ cm fabric samples using several magnets into a metallic panel, which requires little to no expertise and can be done very fast. Pin location does not need to be centimeter-accurate. Because we use depth images, no special lighting is necessary. We measure fabric area density by weighing a $10\times10$ cm sample and dividing by its area.
See Figure~\ref{fig:simScenes} for a visualization of our capture setup and the accompanying video for an illustration of the process.
We capture ten fabrics of diverse compositions and structures, for which we have ground truth mechanical parameters previously obtained with specific equipment and methods~\cite{sperl2022estimation}.

In the supplementary material, we include further details of this dataset, including each material's composition, structure, and closeups.

\subsection{Data Augmentation}
To make our model robust to potential noise and scenario variations common in uncontrolled capture setups (e.g., slightly different camera viewpoint or fabric configuration), we apply several data augmentation strategies to our \emph{synthetic} dataset.

\paragraph*{Simulation-Space Data Augmentation.}
To enforce robustness to different camera viewpoints, simulated meshes are rendered within a range of different inclinations with respect to the vertical plane.

This range covers $\pm5$ degrees from the rest position orientation, creating 11 depth maps per material and scene.  
\paragraph*{Image-Space Data Augmentation.} Real images are largely different to synthetic images, due to distortions, noise, perspective changes, unknown illumination, lens and sensor characteristics, blurs, etc.
To enforce the robustness of our models to those distortions, 
we design an extensive image data augmentation policy consisting of random individual deformations, performed in a particular order.
These not only include random noise, blurs, perspective changes and rescales, but also more complex policies such as thin-plate deformations, posterization and erasing.
This data augmentation policy bridges the gap between synthetic renders and real depth images, which are typically more noisy.
See supplementary material for more details.

%% file: sections_cc/method.tex
\section{Fabric Mechanics from Depth Images}\label{sec:method}

In this section, we present our learning-based approach to estimate fabric mechanical parameters $\hat{\mathcal{P}}$ from depth images. 
Given a set of depth images, $\mathcal{I} = \{ I_{\text{hanging}} \mid I_{\text{stretch}} \} \geq 1$, which depicts the \textit{hanging} and \textit{stretch} scenes,
we train a model $\mathcal{M}$ which maps $\mathcal{I}$, along with the material \textit{density} $\Density$, to mechanical parameters: $\mathcal{M}(\mathcal{I}, \Density) = \hat{\mathcal{P}} \in \mathbb{R}^6$.
To this end, at train time we learn to extract relevant features from depth images, which are then fed into a regressor to learn to predict mechanical features. 
Importantly, our architecture enables to input sets of images at test time by fusing their respective features.
Figure~\ref{fig:model_design} illustrates the train and test pipelines. See the supplementary material for implementation details.

\subsection{Neural Network Architecture}\label{sec:neuralarchitecture}
\begin{figure*}[ht!] 
	\centering
	\includegraphics[width=\textwidth]{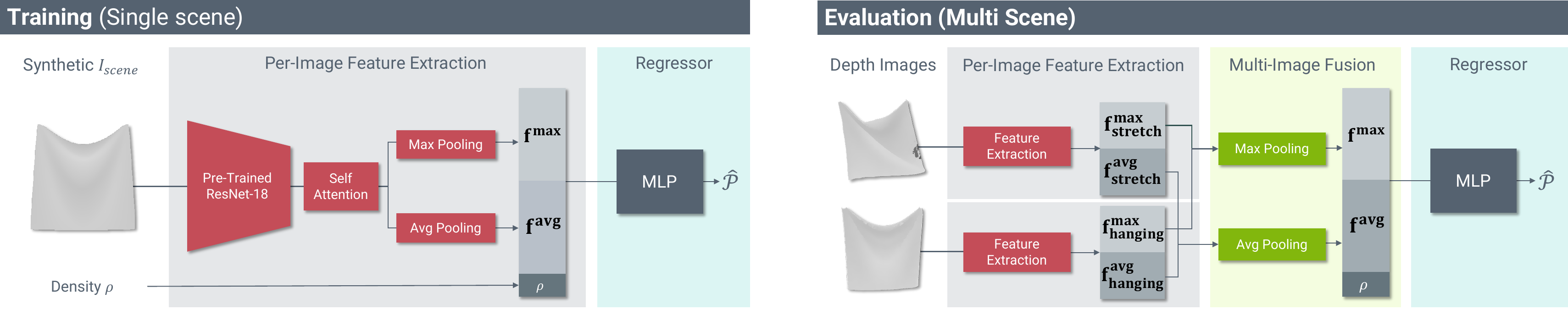}
	\vspace{-5mm}
	\caption[width=1.0\textwidth]{
		Diagram of our training and evaluation pipelines. 
		For \textbf{training}, we use a single image of the material, along with its density. The image is processed by our \emph{Feature Extractor}, followed by an MLP, which computes the parameter estimation $\hat{\mathcal{P}}$. For \textbf{evaluation}, we process each available image with our trained feature extractor and use a \emph{fusion} operator before feeding it to the trained regressor. We use the same Feature Extractor and MLP for both scenes. 
  \vspace{-5mm}
		}
	\label{fig:model_design}

\end{figure*}

\paragraph*{Feature Extractor}
The first part of the model is a \emph{feature extractor} $\mathcal{F}$, which receives as input a single depth image $\dscene \in \mathcal{I}$ and outputs a feature vector $\vect{f}_\text{sc} = \mathcal{F} (\dscene)$ that describes it. This feature extractor is composed of three different components, shown in Figure~\ref{fig:model_design}. 
First, the image is processed by a Convolutional Neural Network (CNN) that outputs a dense latent representation. We use a ResNet-18~\cite{he2016deep} pre-trained on ImageNet~\cite{deng2009imagenet}, which we finetune with our data. 
To reduce the gap between real and synthetic images, we equalize the image before feeding it to the feature extractor, during both training and evaluation.
Then, the output of this CNN is passed through a \emph{Self-Attention} \cite{zhang2019self} module, which  
helps the model learn non-local dependencies, enlarging the model receptive field, so it accounts for distant information in the input images. Self-Attention mechanisms were originally designed for language models~\cite{vaswani2017attention} but have recently demonstrated significant efficacy for computer vision tasks~\cite{ramachandran2019stand, zhang2019self,zhao2020exploring, han2022survey}.
We add a single Self-Attention layer, as they are expensive to train and evaluate. 

Finally, we perform pooling operations to transform the output of the Self-Attention layer to a feature vector, $\vect{f}_\text{sc}$, of a fixed size.
In addition to the commonly used max-pooling~\cite{simonyan2014very}, we further concatenate it with the output of average-pooling, which has shown  to improve performance of attention modules~\cite{zhou2016learning,hu2018squeeze, woo2018cbam}. The feature vector $\vect{f}_\text{sc}$ is thus a concatenation of max-pooled features and average-pooled features: $\vect{f}_\text{sc} =  \{\vect{f}^{max}_\text{sc} \oplus \vect{f}^{avg}_\text{sc}\}$. 

\paragraph*{Fusion} 
Our design allows us to combine features $\vect{f}_\text{sc}$ from more than one scene into a single feature vector $\vect{f}$.
For every image in $I$, we compute $\vect{f}_\text{sc}$. We then \emph{fuse} those feature vectors into a single vector by performing pooling across $\vect{f}$. Similarly to $\vect{f}_\text{sc}$, $\vect{f}$ is composed of two types of features: $\vect{f} = \{max_{sc} \{\vect{f}^{max}_{sc}\} \oplus \{avg_{sc} \{\vect{f}^{avg}_{sc}\}\}$. 
The max-pooled features are fused using the maximum value across scenes, while the average-pooled features are fused using the mean across scenes. We illustrate this procedure in Figure~\ref{fig:model_design}.

\paragraph*{Parameter Regressor}
 Our last component is a fully-connected Multi-Layer Perceptron (MLP), which takes as input the feature vector $\vect{f}$ and the material density, $\Density$, and outputs the simulation parameters $\hat{\mathcal{P}}$. 
 As our loss function, we compare the real parameters $\mathcal{P}$ with the model estimations $\mathcal{M}(\mathcal{I}, \Density) = \hat{\mathcal{P}}$ using an $\ell_2$ norm.

%% file: sections_cc/evaluation_params.tex
\subsection{Quantitative Evaluation}
\label{sec:evaluation_params}

In this section we quantitatively evaluate the performance of the method for estimating the mechanical parameters. 
We validate the design choices of the model, and evaluate our results depending on the type of input used.

\begin{table}[tb]
	\centering
	\resizebox{\columnwidth}{!}{
		\begin{tabular}{@{}rlccccccc@{}}
			\toprule
			&
			\multicolumn{1}{l}{} &
			\multicolumn{1}{c}{\textbf{Baseline}} &
			\multicolumn{2}{c}{\textbf{Data Augmentation}} &
			\multicolumn{3}{c}{\textbf{Architecture}} \\ \cline{4-9} 
			\textbf{Metric} &
			\multicolumn{1}{l}{\textbf{Parameter}} &
			\multicolumn{1}{l|}{\textbf{}} &
			\multicolumn{1}{r}{\textbf{w/ sim}} &
			\textbf{w/ image} &
			\multicolumn{1}{|l}{\textbf{w/ pre-Train}} &
			\textbf{w/ attention} &
			\textbf{w/ pooling} \\ \hline
			\multirow{8}{*}{$\ell_1 \downarrow$} &
			\multicolumn{1}{l|}{kStretchWeft} & \multicolumn{1}{r|}{\RedColor{0.122}} &
			0.118 &
			\multicolumn{1}{r|}{0.112} &
			0.087 &
			
			\GreenColor{0.071} &
			\GreenColor{0.071}  \\
			&
			\multicolumn{1}{l|}{kStretchWarp} & \multicolumn{1}{r|}{\RedColor{0.109}} &
			0.101 &
			\multicolumn{1}{r|}{0.102} &
			{0.074} &
			
			0.066 &
			\GreenColor{0.061}  \\
			&
			\multicolumn{1}{l|}{kStretchBias} & \multicolumn{1}{r|}{\RedColor{0.050}} &
			0.043 &
			\multicolumn{1}{r|}{0.046} &
			{{0.043}} &
			0.042 &
			\GreenColor{0.038} &\\
			&
			\multicolumn{1}{l|}{\textbf{Avg. Stretch}} & \multicolumn{1}{r|}{\RedColor{0.094}} &
			0.087 &
			\multicolumn{1}{r|}{0.087} &
			{{0.068}} &
			0.060 &
			\GreenColor{0.057} & \\ \cline{2-8} 
			&
			\multicolumn{1}{l|}{kBendingWeft} & \multicolumn{1}{r|}{\RedColor{0.095}} &
			0.093 &
			\multicolumn{1}{r|}{0.091} &
			{{0.084}} &
			0.0.82&
			\GreenColor{0.072} & \\
			&
			\multicolumn{1}{l|}{kBendingWarp} & \multicolumn{1}{r|}{\RedColor{0.124}} &
			0.119 &
			\multicolumn{1}{r|}{0.112} &
			{0.110} &
			0.101 &
			\GreenColor{0.094} & \\
			&
			\multicolumn{1}{l|}{kBendingBias} & \multicolumn{1}{r|}{\RedColor{0.086}} &
			0.081 &
			\multicolumn{1}{r|}{\RedColor{0.086}} &
			0.081 &
			0.068 &
			\GreenColor{0.060} & \\
			&
			\multicolumn{1}{l|}{\textbf{Avg. Bending}} &\multicolumn{1}{r|}{\RedColor{0.102}} &
			0.098 &
			\multicolumn{1}{r|}{0.097} &
			{{0.092}} &
			0.084 &
			\GreenColor{0.075}  \\ \hline
			\multirow{7}{*}{$r \uparrow$} &
			\multicolumn{1}{l|}{kStretchWeft} & \multicolumn{1}{r|}{\RedColor{0.445}} &
			0.475 &
			\multicolumn{1}{r|}{0.453} &
			{{0.643}} &
			0.788 &
			\GreenColor{0.798}
			\\
			&
			\multicolumn{1}{l|}{kStretchWarp} & \multicolumn{1}{r|}{{0.543}} &
			\RedColor{0.503} &
			\multicolumn{1}{r|}{0.510} &
			{{0.645}} &
			0.715 &
			\GreenColor{0.771} &
			\\
			&
			\multicolumn{1}{l|}{kStretchBias} & \multicolumn{1}{r|}{{\RedColor{0.477}}} &
			{0.508} &
			\multicolumn{1}{r|}{0.608} &
			0.701 &
			0.733 &
			\GreenColor{0.781} &
			\\
			&
			\multicolumn{1}{l|}{\textbf{Avg. Stretch}} & \multicolumn{1}{r|}{\RedColor{0.488}} &
			0.495 &
			\multicolumn{1}{r|}{0.523} &
			0.660 &
			0.745 &
			\GreenColor{0.783} &
			\\ \cline{2-8} 
			&
			\multicolumn{1}{l|}{kBendingWeft} & \multicolumn{1}{r|}{\RedColor{0.611}} &
			0.684 &
			\multicolumn{1}{r|}{0.781} &
			0.783 &
			0.798 &
			\GreenColor{0.863} &
			\\
			&
			\multicolumn{1}{l|}{kBendingWarp} & \multicolumn{1}{r|}{\RedColor{0.614}} &
			0.654 &
			\multicolumn{1}{r|}{0.747} &
			{{0.772}} &
			0.806 &
			\GreenColor{0.921} &
			\\
			&
			\multicolumn{1}{l|}{kBendingBias} & \multicolumn{1}{r|}{\RedColor{0.624}}&
			0.828 &
			\multicolumn{1}{r|}{0.837} &
			0.872 &
			0.893 &
			\GreenColor{0.942} &
			\\
			&
			\multicolumn{1}{l|}{\textbf{Avg. Bending}} & \multicolumn{1}{r|}{\RedColor{0.616}} &
			0.722 &
			\multicolumn{1}{r|}{0.788} &
			{{0.809}} &
			{0.832} &
			\GreenColor{0.909} &
			\\ \cline{1-8} 
	\end{tabular}}
	\caption{Ablation study of the neural architecture and data augmentation. 
		From left to right, we build upon our baseline and progressively add: simulation-space data augmentation, image-space data augmentation, pre-training, self-attention, and average-pooling. 
		On both MAE ($\ell_1$) and correlation ($r$) metrics, we observe increased performance on the validation set in every added component.
		Using a pre-trained network for feature extraction yields the largest gains.
		We use a color code to highlight \GreenColor{best} and \RedColor{worst} cases. }
	\label{tab:ablation}
\end{table}

\subsubsection{Ablation Study of the Model Design}
\label{sec:modeldesign}

We aim to understand the effective contribution of the data augmentation strategy, and the network architecture design.
For these experiments, we randomly split the synthetic data in $90\%$ for training and $10\%$ for validation, using the same split for every experiment.
Results are shown in Table~\ref{tab:ablation}. 
As our baseline, we use a simple model without \textit{self-attention}, without \textit{average-pooling} features, where the CNN backbone is randomly initialized, and without data augmentation. From this baseline, we progressively add different components and measure the performance of the validation data using Mean Absolute Error (MAE) and Spearman correlation ($r$). The parameters are normalized using the minimum and maximum values of the training set.

Given the training configuration with all the data augmentation --which provides a small increase in performance  most likely because the validation dataset is synthetic data-- we evaluate the neural architecture. 

Using a CNN backbone pre-trained on ImageNet~\cite{deng2009imagenet} instead of a randomly initialized one, we observe a significant increase in model performance across every parameter and metric. Training a feature extractor that receives images from both scenes at the same time would not allow us to leverage pre-training, which would negatively impact generalization.
Then, we add a \textit{self-attention}~\cite{zhang2019self} layer after the CNN backbone, which allows the model to integrate information that is present on distant areas of the images. 
Interestingly, this module significantly helps predict the \emph{kStretch} parameters while having a more minor influence on the \emph{kBending}.
Finally, adding \textit{average-pooling} in addition to the commonly used \textit{max-pooling} have a highly positive impact in the error rates. We use this last configuration with all the components for all the results shown in the paper. It is worth noting that the \emph{kBendingBias} is easily predicted by the model, even in its most basic configuration. 
This is likely because this parameter correlates most strongly with the density of the material (shown in Figure~\ref{fig:dataset-statistics}), so the model can leverage this information for the predictions.

\begin{table}[tb]
	\centering
	\resizebox{\columnwidth}{!}{
		\begin{tabular}{@{}rlrrrrrrr@{}}
			\toprule
			&
			\multicolumn{1}{l}{} &
			\multicolumn{1}{c}{\textbf{Density}} &
			\multicolumn{3}{c}{\textbf{Only Depth}} &
			\multicolumn{3}{c}{\textbf{Density \& Depth}} \\ \cline{4-9} 
			\textbf{Metric} &
			\multicolumn{1}{l}{\textbf{Parameter}} &
			\multicolumn{1}{l|}{\textbf{}} &
			\multicolumn{1}{l}{\textbf{Stretch}} &
			\textbf{Hanging} &
			\textbf{Both} &
			\multicolumn{1}{|r}{\textbf{Stretch}} &
			\textbf{Hanging} &
			\textbf{Both} \\ \hline
			\multirow{8}{*}{$\ell_1$ $\downarrow$} &
			\multicolumn{1}{l|}{kStretchWeft} & \multicolumn{1}{r|}{\RedColor{0.113}} &
			0.102 &
			0.107 &
			\multicolumn{1}{r|}{0.062} &
			0.054 &
			{0.056} &
			\GreenColor{0.051} \\
			&
			\multicolumn{1}{l|}{kStretchWarp} & \multicolumn{1}{r|}{\RedColor{0.091}} &
			0.067 &
			{0.081} &
			\multicolumn{1}{r|}{{0.056}} &
			0.055 &
			{0.059} &
			\GreenColor{0.052} \\
			&
			\multicolumn{1}{l|}{kStretchBias} & \multicolumn{1}{r|}{0.034} &
			0.039 &
			\RedColor{0.054} &
			\multicolumn{1}{r|}{{0.034}} &
			0.033 &
			{0.036} &
			\GreenColor{0.031} \\
			&
			\multicolumn{1}{l|}{\textbf{Mean Stretch}} & \multicolumn{1}{r|}{0.079} &
			0.069 &
			\RedColor{0.081} &
			\multicolumn{1}{r|}{{0.051}} &
			0.047 &
			{0.050} &
			\GreenColor{0.045} \\ \cline{2-9} 
			&
			\multicolumn{1}{l|}{kBendingWeft} & \multicolumn{1}{r|}{0.142} &
			\RedColor{0.233} &
			{0.213} &
			\multicolumn{1}{r|}{{0.145}} &
			0.128 &
			{0.139} &
			\GreenColor{0.125} \\
			&
			\multicolumn{1}{l|}{kBendingWarp} & \multicolumn{1}{r|}{0.126} &
			\RedColor{0.184} &
			{0.126} &
			\multicolumn{1}{r|}{{0.074}} &
			0.037 &
			{0.063} &
			\GreenColor{0.035} \\
			&
			\multicolumn{1}{l|}{kBendingBias} & \multicolumn{1}{r|}{0.094} &
			\RedColor{0.166} &
			0.081 &
			\multicolumn{1}{r|}{{0.054}} &
			0.055 &
			{0.069} &
			\GreenColor{0.046} \\
			&
			\multicolumn{1}{c|}{\textbf{Mean Bending}} &\multicolumn{1}{r|}{0.121} &
			\RedColor{0.194} &
			{0.140} &
			\multicolumn{1}{r|}{{0.091}} &
			0.073 &
			{0.090} &
			\GreenColor{0.069} \\ \hline
			\multirow{8}{*}{ $r \uparrow$} &
			\multicolumn{1}{l|}{kStretchWeft} & \multicolumn{1}{r|}{\RedColor{0.184}} &
			0.407 &
			{0.403} &
			\multicolumn{1}{r|}{{0.418}} &
			0.712 &
			{0.469} &
			\GreenColor{0.728} \\
			&
			\multicolumn{1}{l|}{kStretchWarp} & \multicolumn{1}{r|}{\RedColor{0.002}} &
			0.502 &
			{0.407} &
			\multicolumn{1}{r|}{{0.503}} &
			0.520 &
			{0.433} &
			\GreenColor{0.533} \\
			&
			\multicolumn{1}{l|}{kStretchBias} & \multicolumn{1}{r|}{0.289} &
			\RedColor{-0.141} &
			-0.068 &
			\multicolumn{1}{r|}{{-0.007}} &
			{0.367} &
			0.383 &
			\GreenColor{0.550} \\
			&
			\multicolumn{1}{l|}{\textbf{Mean Stretch}} & \multicolumn{1}{r|}{\RedColor{0.158}} &
			0.256 &
			\RedColor{0.247} &
			\multicolumn{1}{r|}{{0.305}} &
			0.533 &
			{0.428} &
			\GreenColor{0.604} \\ \cline{2-9} 
			&
			\multicolumn{1}{l|}{kBendingWeft} & \multicolumn{1}{r|}{0.483} &
			\RedColor{0.052} &
			0.267 &
			\multicolumn{1}{r|}{{0.625}} &
			{0.673} &
			0.683 &
			\GreenColor{0.717} \\
			&
			\multicolumn{1}{l|}{kBendingWarp} & \multicolumn{1}{r|}{0.317} &
			\RedColor{0.048} &
			{0.156} &
			\multicolumn{1}{r|}{{0.407}} &
			0.467 &
			{0.433} &
			\GreenColor{0.533} \\
			&
			\multicolumn{1}{l|}{kBendingBias} & \multicolumn{1}{r|}{0.357}&
			\RedColor{-0.044} &
			0.108 &
			\multicolumn{1}{r|}{{0.250}} &
			{0.333} &
			0.417 &
			\GreenColor{0.546} \\
			&
			\multicolumn{1}{l|}{\textbf{Mean Bending}} & \multicolumn{1}{r|}{0.386} &
			\RedColor{0.019} &
			0.177 &
			\multicolumn{1}{r|}{{0.427}} &
			{0.491} &
			0.511 &
			\GreenColor{0.599} \\ \cline{1-9} 
	\end{tabular}}
	\caption{Results for real depth images varying the input. From left to right, the input is: only density, only depth images, and both density and depth. 
	The best results are obtained using every data source as input. For \emph{kBending}, the \emph{density} alone provides more information than only depth images. 
		We use a color code to highlight \GreenColor{best} and \RedColor{worst} cases. }
	\label{tab:one_vs_multiple}
\end{table}

\subsubsection{Evaluation of Input Influence }\label{sec:evalreal}

The design of our method supports taking as input one or multiple images. In this experiment, presented in Table~\ref{tab:one_vs_multiple}, we evaluate the error testing different configurations of the input. Note that we train a different model for each configuration.
In every case, using the \emph{density} as input helps the model to generalize. This is particularly relevant for \emph{kBending} parameters, for which the \emph{density} alone provides more information than depth images. The \emph{stretch} scene is typically more informative than the \emph{bending} one, as both MAE and correlations are usually better when it is provided. When using both scenes simultaneously, the model provides more accurate estimations than any of the scenes individually, showing that the two scenes provide complimentary information.

\subsubsection{Neural Saliency Maps}
\label{sec:saliency}

\begin{figure}[htb]
	\centering
	
	\includegraphics[width=\columnwidth]{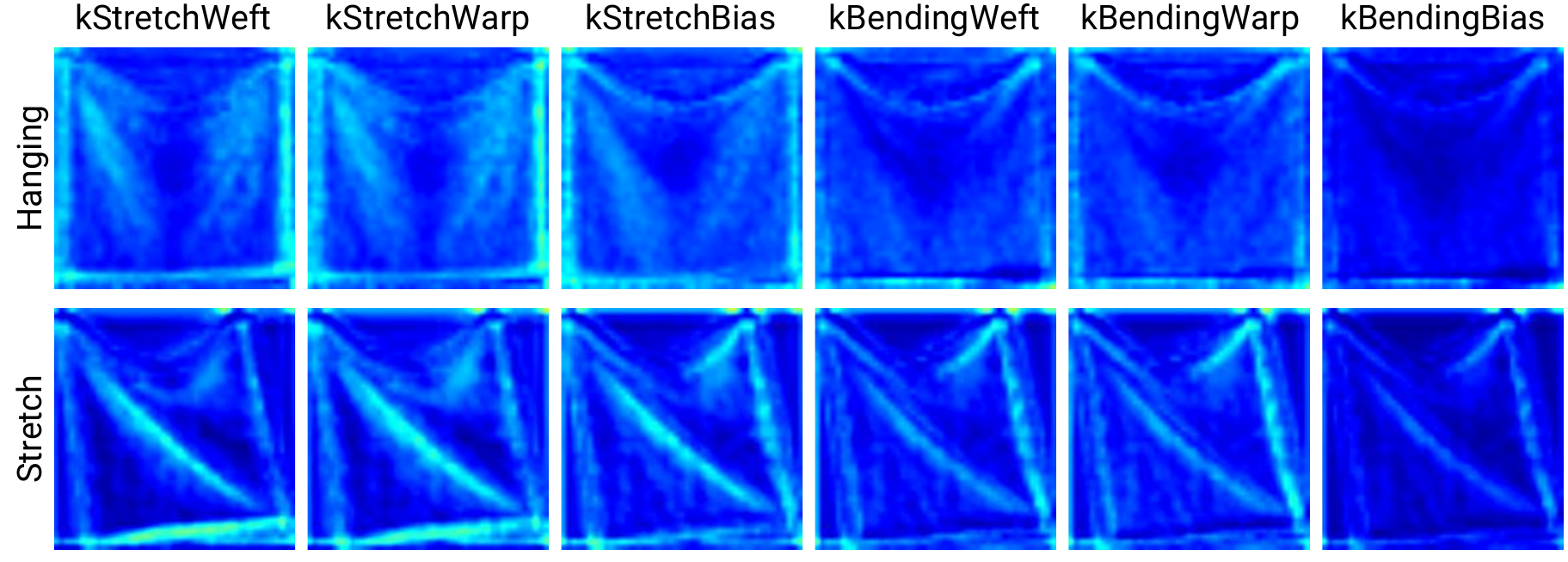}
	\caption[width=1.0\columnwidth]{\label{fig:saliency}{Saliency maps~\cite{FuHDGGL20} aggregated per parameter.
			The model relies on the central areas of the fabric samples for predicting the stretch parameters. For bending, it is most sensitive to areas on the borders of the samples. }
		\vspace{-3mm}
	}

\end{figure}

We aim to understand which part of the scenes are most relevant for the model when making its predictions. To do so, we create saliency maps using \emph{Axion-Based Class-Activation Mappings}~\cite{FuHDGGL20}, that averages the activations of the deep layers weighted by their importance with respect to each target parameter, and aggregate them over our real dataset. 
In Figure~\ref{fig:saliency} we observe that each scene provides the model with different cues. For \emph{kStretch} parameters, the model is sensitive to the central wrinkle of the \emph{stretch} scene and the central fold of the \emph{hanging}. 
For \emph{kBending} parameters, the model is sensitive to areas in the borders of the fabric where small but noticeable wrinkles are present, in both scenes.
As in other experiments, we observe that the model can find more relevant features on the \emph{stretch} scene.

%% file: sections_cc/distancemetric.tex
\section{A Similarity Metric for Drape}
\label{sec:distance_metric}

The regressor introduced in Section \ref{sec:method} allows to infer the mechanical parameters of target fabric.
However, since a direct interpretation of such parametric space is not human friendly, it is nearly impossible to understand the residual errors shown in Tables \ref{tab:ablation} and \ref{tab:one_vs_multiple}.
Do the regressed parameters produce a drape similar to the target image?
Notice that, since the mechanical parameter space is non-orthogonal, small parameter changes may produce unexpected deformations.  
Therefore, we hypothesize that a \textit{perceptual} similarity metric for drape is needed to interpret our quantitative results. We describe this metric next.

\subsection{Image-based Similarity of Drape} \label{sec:image_distance_metric}
Motivated by our hypothesis that the \textit{hanging} and \textit{stretch} scenes are sufficient to convey the fabric mechanics, we propose an image-based similarity metric using renders of such scenes. 
Let $\mathcal{P} \subseteq \mathbb{R}^7$ be the parameter space of our simulator, $P_a \in \mathcal{P}$ and $P_b \in \mathcal{P}$ two different parameter sets, and $a \sim \mathcal{R} (P_a , \text{scene})$ and $b \sim \mathcal{R} (P_b, \text{scene})$ two rendered simulations obtained for a certain scene configuration. 
We define a distance metric for a particular scene as 
\begin{equation}\label{eq:metric}
	d_{\text{scene}} (P_a, P_b) = \frac{\sum^{N}_{i=1} \sum^{N}_{j=1} \text{IM}(a_i, b_j)}{N^2}
\end{equation}
where $\text{IM}$ is an image-space distance metric, and $N$ is the number of different simulations we run.
Since real cloth is very sensitive to parameters such as initial state or initial shape, in order to learn a metric that is robust to real-world conditions, we perturb the initial state and boundary conditions in a set of simulations using random jittering to the initial forces. 
We empirically found that averaging over multiple simulations for the same set of $(P_a,P_b)$ gives us a more informative metric.

Further more, we take into account both \textit{hanging} and \textit{stretch} scenes, hence our final metric is defined by averaging their distances across both scenarios, resulting in our final metric:
\begin{equation}
	d (P_a, P_b)  = \frac{d_{\text{hanging}} (P_a, P_b)  + d_{\text{stretch}}  (P_a, P_b)}{2}
\end{equation}

Note that we propose a similarity metric, which, as opposed to real distance metrics, does not necessarily have to meet the metric axioms~\cite{tversky1974judgment}: it can produce asymmetric values, violate the triangle inequality, and does not need to define what the identity is. 
According to the Equation~\ref{eq:metric}, the distance of one fabric with itself is not necessarily zero; it just needs to satisfy a minimum requirement where the distance of every material with itself should be smaller than the distance of any material with any other material,
\begin{equation}
	d (P_a, P_a)  < d (P_a, P_b), \forall P_a \neq P_b
\end{equation}
In order to remove the possible influence of optical properties, scene illumination, and camera parameters, we use the same scene configuration for every render, with grayscale albedo and a lambertian BRDF.

For $\text{IM}$ we use LPIPS~\cite{zhang2018unreasonable} which we empirically found to perform better than other alternative image metrics. 
We find that metrics based on pre-trained neural networks work better than lower-level alternatives, while content-aware distances are more powerful for this purpose than style-aware metrics.
This suggests that the size, position and shape of the wrinkles and deformations of the fabrics are important factors that explain differences between materials. 
We empirically found that $N=5$ simulations is typically enough, as more samples provide very marginal improvements. 
See the supplementary material for more details about the proposed metric.

%% file: sections_cc/results.tex
\section{Evaluation}

We propose to evaluate our method from Section \ref{sec:method} and the metric from Section \ref{sec:distance_metric} by comparing our estimations with human ratings (recall that we provide quantitative errors per parameter in Section~\ref{sec:evaluation_params}).
To this end, in Section \ref{sec:humanjudgements}, we first collect a large number of ground truth human judgments about the similarity
of triplets of fabrics. 
Then, in Section \ref{sec:image-vs-human}, we demonstrate that our image-based metric using ground truth parameters, as well as the estimated ones, encode the same preferences.
Finally, in Section \ref{sec:qualitativeresults}, we show qualitative comparisons and demonstrate the usefulness of our approach in a downstream task consisting of `search by similarity'.

\subsection{Human Judgment Perceptual Similarity of Drape}
\label{sec:humanjudgements}

\begin{figure}[tb]
	\centering
	\includegraphics[width=0.9\columnwidth]{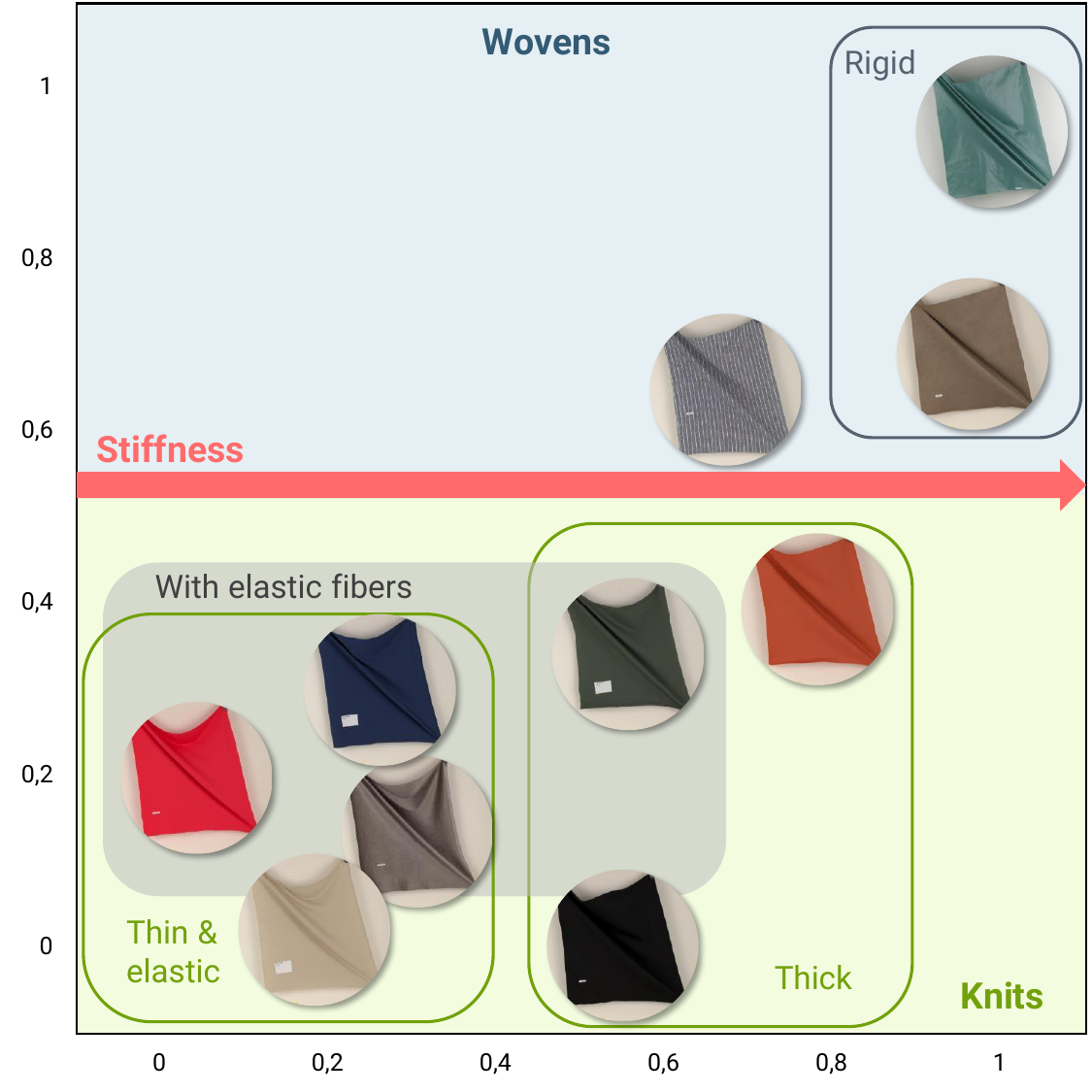}
	\vspace{-4mm}
	\caption{Human Judgments 2D tSTE embedding ~\cite{tamuz2011adaptively} computed from human perceptual judgments about real fabric similarity (i.e., ground truth).
	We observe interesting patterns: woven and knits are separated; elastic materials are clustered; thick and thin materials are separated. Neither axes directly correspond to any material property, instead they emerge from the embedding. 
	}
 \vspace{-3mm}
	\label{fig:tste}
\end{figure}

We then use ten samples from our real dataset with known ground truth mechanical parameters and setup the user study as follows.
Participants are presented with a triplet of fabrics and, using one fabric as a reference, they are asked which of the two remaining fabrics is most similar \cite{zhang2018unreasonable,garces2014similarity} to the reference fabric.
Participants are encouraged to manipulate the samples and focus only on the mechanical similarity and overall drape, and to ignore properties like material reflectance.
Each participant rated 20 triplets that were
pseudo-randomly sampled, ensuring that at least each of the ten test fabrics is used twice as reference. 
Given the same triplet, we observe an average of $86.68\%$ agreement between our participants across all experiments and materials, suggesting that there is a perceptual understanding of fabrics mechanics that humans share.
We did not observe any significant differences in agreement depending on the volunteer demographics or level of expertise in fabric handling or simulation. 

Leveraging the user study described above, we can compute an embedding that captures the relative distance between real materials according to human perception (i.e., ground truth perceptual similarity).
Figure~\ref{fig:tste} depicts such embedding in 2D, computed using tSTE \cite{tamuz2011adaptively}, which allows us to calculate perceptual distances between materials using the Euclidean norm.
The embedding depicts many interesting patterns, including perfectly separated woven and knitted fabrics; elastic and thin fabrics are clustered together; thick and thin knits are well separated.
These patterns suggest that fabric structure, composition, and density play an important, non-linear role in the overall perception of the mechanical properties of fabrics.
\begin{figure}[t]
	\includegraphics[width=\columnwidth]{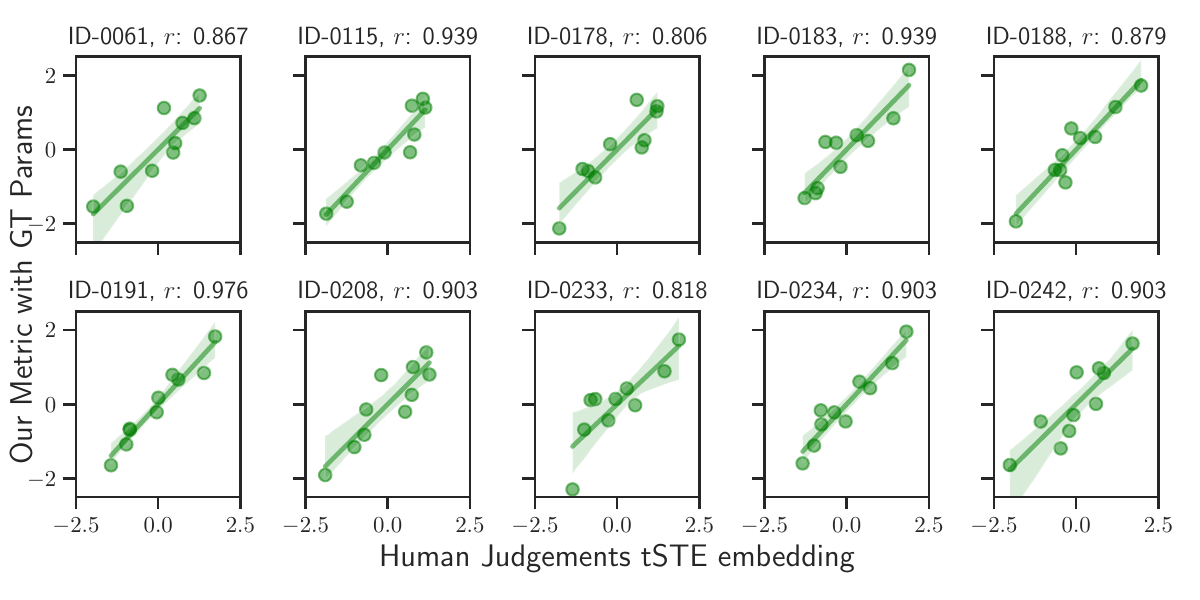}
	\includegraphics[width=\columnwidth]{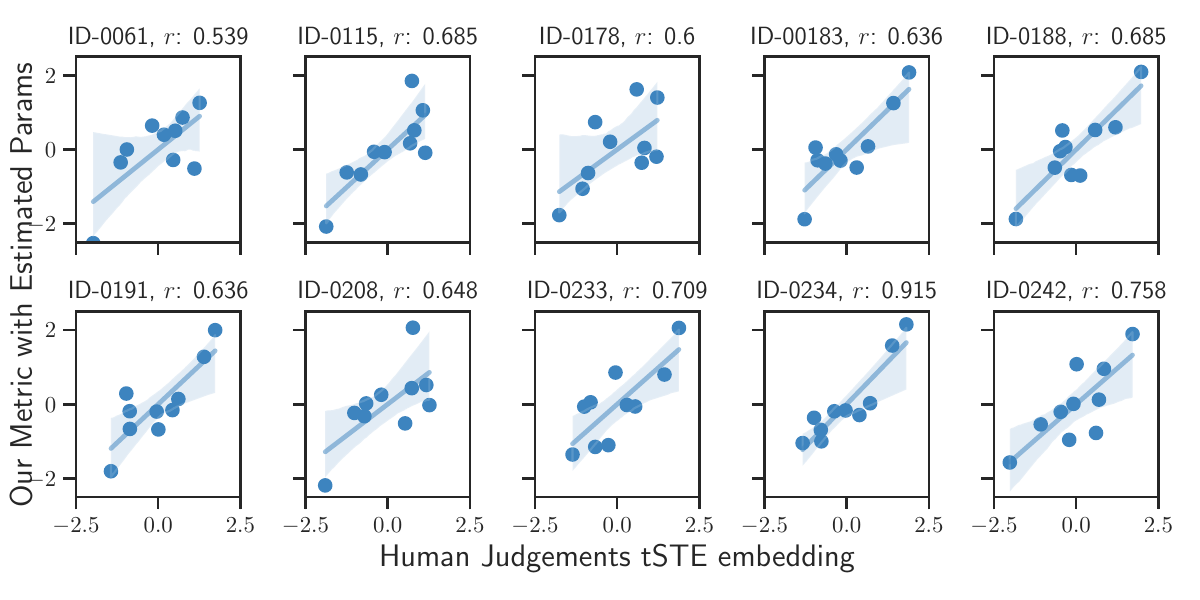}
	\caption{Correlation between the ordering provided by the Human Judgments (x-axis) and our drape similarity metric with (top) the Ground Truth simulation parameters (y-axis), and (bottom) the estimations of our model. 
		We plot z-scores instead of the raw distances to help visualization.}
	\label{fig:HUMANS_VS_METRIC_v2}
\end{figure}

\subsection{Image-based vs. Human Perceptual Similarity}
\label{sec:image-vs-human}

We evaluate the agreement between humans and our estimations within the context of a similarity rank. For each fabric of our real dataset, we compute the distance to the rest of the materials using different metrics: 1) the Euclidean distances on the Human Judgments tSTE embedding shown Section~\ref{sec:humanjudgements}; 2) the z-score distances in the parametric space of the mechanical simulation, $\mathcal{P}$; 3) the distance using our similarity metric for drape explained in Section~\ref{sec:distance_metric}. We compare ground truth parameters and estimated ones for the second and third cases. 
The summary of results is shown in Table~\ref{tab:distances_vs_humans}, and the complete analysis is presented in the supplementary material.

First, we demonstrate that our similarity metric using ground truth parameters correlates with human judgments. Figure~\ref{fig:HUMANS_VS_METRIC_v2}~(top) illustrates the outcome using Spearman correlation ($\textrm{r}$). We observe that the correlation for each fabric is higher than $0.8$, with an average of $0.893$, showcasing a strong correlation. These results suggest that our metric, which only takes images of the \emph{hanging} and \emph{stretch} scenes as input, can measure distances between materials as humans would.
Then, as shown in Figure~\ref{fig:HUMANS_VS_METRIC_v2}~(bottom), we use our estimated parameters instead of the ground truth, reaching an average correlation of $0.680$. Even though this value is slightly smaller than the ground truth, it is still significant to conclude the estimations of our model agree with human judgments. Note that
the materials with higher correlation are those lying on the extreme
areas of the embedding obtained in Figure~\ref{fig:tste} that have very distinct characteristics.
Likewise, we also compute the distance for each material using merely the parameter spaces of the simulation. As can be seen, using this space does not produce correlated outputs with human judgments, reaching correlations below $0.44$ in any case tested.

\begin{table}[t]
	\centering
	\resizebox{0.8\columnwidth}{!}{
	\begin{tabular}{@{}rcc@{}}
		\cmidrule(l){2-3}
		\multicolumn{1}{l}{}          & \textbf{Parameter Distance} & \textbf{Similarity Metric} \\ \midrule
		\textbf{GT}        & $0.431\pm0.17$              & \GreenColor{\textbf{0.893}}$\pm0.05$            \\
		\textbf{Estimated} & $0.377\pm0.19$              & \GreenColor{\textbf{0.680}}$\pm0.09$            \\ \bottomrule
	\end{tabular}}
	\caption{Average ($\pm$ std.) correlation with rankings obtained through the Human Judgments tSTE Embedding, depending on the parameter source (ground truth or predicted), and metric used to compute similarity (parameter distance or our drape similarity).}
	\vspace{-3mm}
	\label{tab:distances_vs_humans}
\end{table}

%% file: sections_cc/resultsqualitative.tex
\subsection{Qualitative Results}\label{sec:qualitativeresults}

\begin{figure}[t]
	\includegraphics[width=\columnwidth]{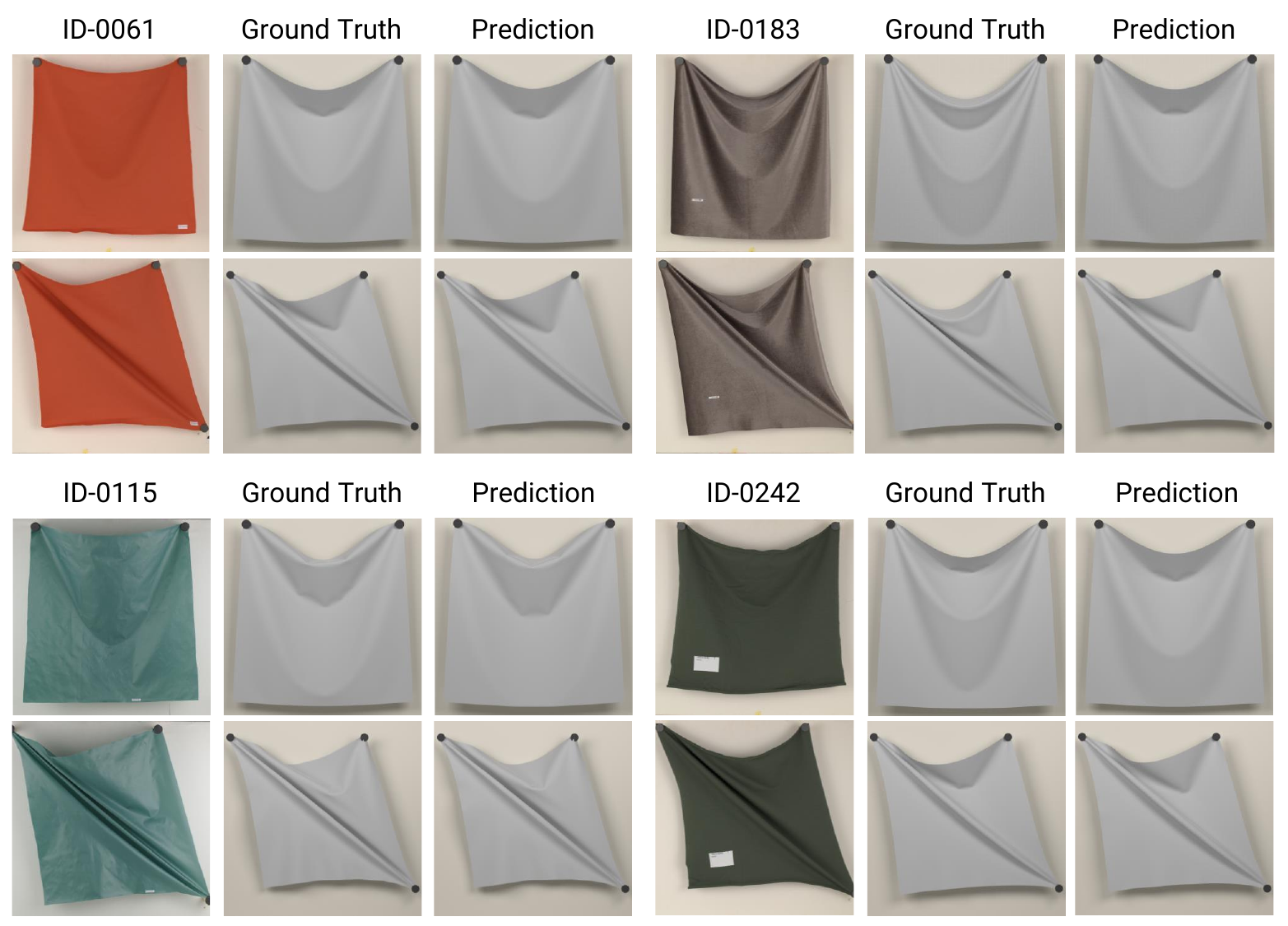}
	\caption{A comparison between the simulations obtained through the ground truth parameters, and those obtained using the predictions of our model, from a representative set of fabrics of our test set. As shown, the estimations of the model yield similar drapes to those of their ground truth counterparts, which we also evaluate quantitatively. }
	\label{fig:qualitative_results}
	\vspace{-3mm}
\end{figure}

\begin{figure}[t]
	\includegraphics[width=\columnwidth]{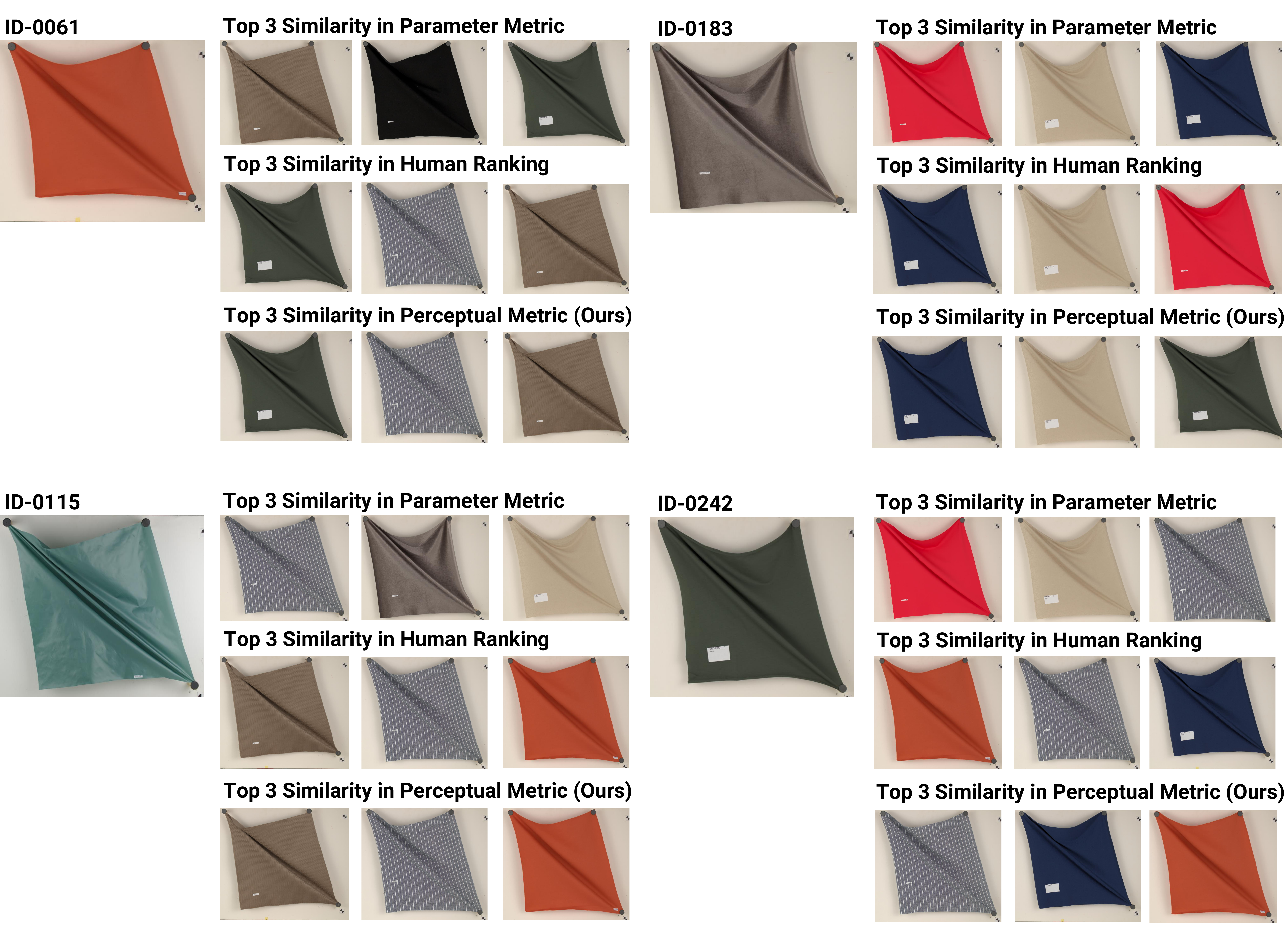}
	\caption{Search by drape similarity. The ordering provided by the parameter space (first row) does not match human judgments (second row), while the arrangement obtained by our metric matches humans with high consensus (third row).  
	}
\vspace{-5mm}
	\label{fig:similarities_small}
\end{figure}

Figure~\ref{fig:qualitative_results} compares the simulations obtained with the parameters of our method with the ground truth parameters for a few examples. We can observe that our estimations are very close to the ground truth in this cases. The full results are contained in the supplementary material.
Finally, our metric can be used to search between materials of similar drape. We illustrate this in Figure~\ref{fig:similarities_small}. As shown, a naive ranking using directly the parameter space does not provide any meaningful ordering. On the contrary, using our similarity metric, we obtain ranks that agree with those given by the human embedding, showcasing the potential of our automatic metric to explore fabric collections.  

\begin{figure}[t]
	\includegraphics[width=\columnwidth]{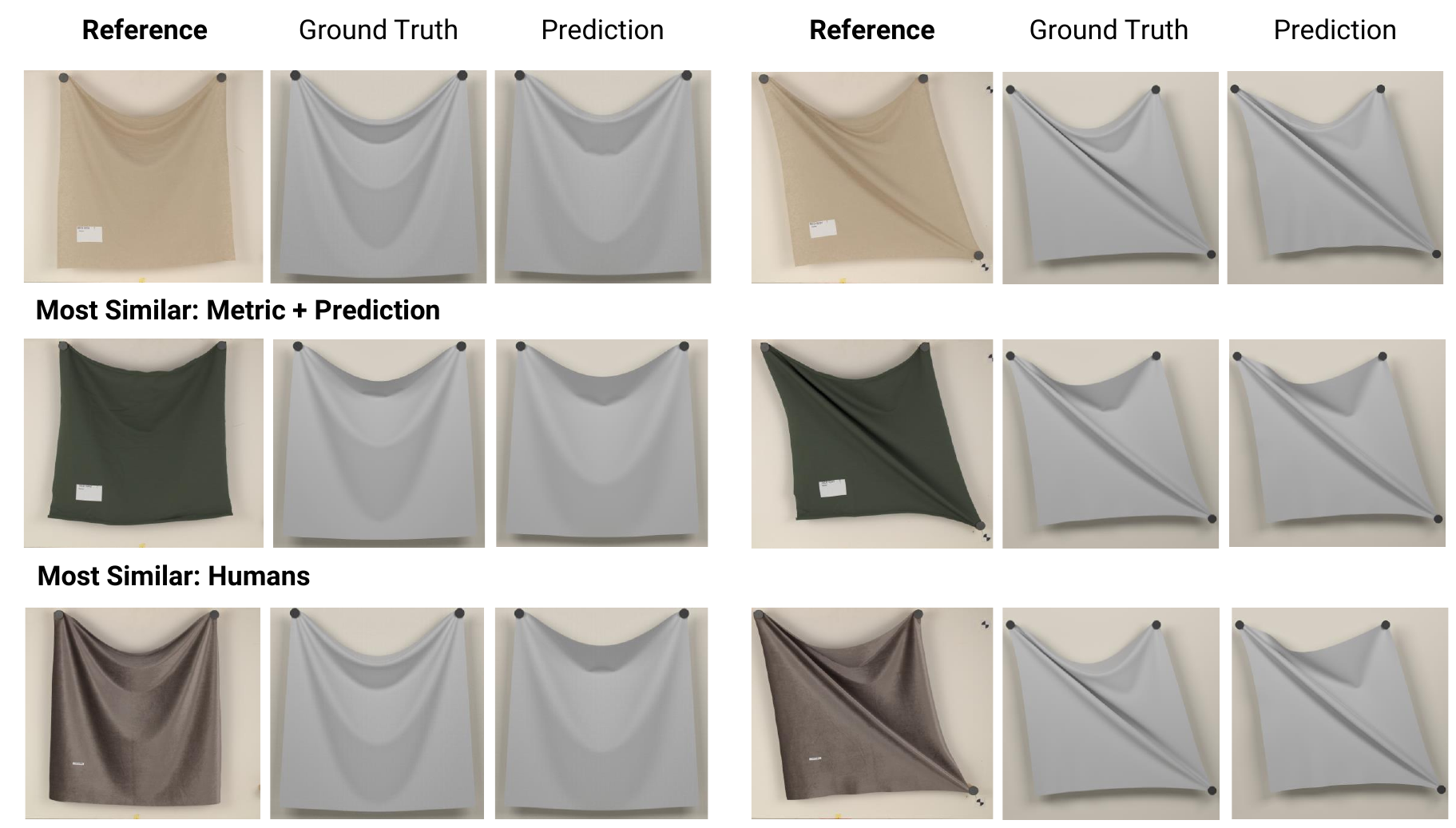}
	\caption{A failure case of our method. Given the reference material (first row) as input, the model predicts fewer bends than the ground truth. According to our metric, this prediction is closer to a thicker material (middle row), than to what humans perceive as most similar to the reference fabric (bottom row).
	}
	\label{fig:failure_case}
\end{figure}

\paragraph*{Limitations} Even if our model provides accurate predictions, its estimations are not always truthful to the real materials. We illustrate this in Figure~\ref{fig:failure_case}, where the model predicts fewer bends on the final drape than what the ground truth generates. According to our metric, this prediction is closer to thicker materials than to what humans perceive as most similar to the reference material.

%% file: sections_cc/conclusions.tex
\section{Conclusions} \label{sec:conclusions}
In this work, we have presented a casual method to estimate mechanical parameters of fabrics from depth images of fabric samples placed at two specific configurations. We have validated our architecture and inputs numerically, proving that all the components of our method are necessary to provide accurate estimations. 
While our quantitative analysis helped us understand the importance of each component, we found that these errors are not interpretable, nor do they help us understand the overall appearance of the predicted drape. Therefore, we have presented the first metric, which, by purely working on the image space, can capture differences in fabric mechanics like humans do. We have used such metric for two purposes: first, to validate the accuracy of our estimated parameters perceptually, and second, to showcase a novel application of search by drape similarity.  

Our work could be improved in several ways. 
Our neural network is trained using a purely regression loss. Training the network using differentiable simulation could improve training, and help generalization and error interpretation. We could incorporate our perceptual metric as a loss function. However, it requires multiple differentiable simulations and a deep feature extractor, which will result in a significant computational overhead. In addition, less expressive simulation engines may correlate less with human perception. Similarly, we would like to scale our training dataset and user study to handle more and more diverse samples, to cover a broader variety of fabric families. Interesting possible extensions would include taking as input RGB images instead of depth maps, training with real samples, incorporating symmetry consistency losses, or learning a similarity metric that can work by using captured images as input (instead of simulations).
Finally, we hope our work might inspire future work in the long-standing problem of validating fabric mechanics in a way that is agnostic to the simulator parameters. 

\paragraph*{Acknowledgments}

We would like to thank Gabriel Cirio and Alejandro Rodríguez for valuable discussions, Alicia Nicas for her help with the simulations, and Sofía Domínguez for her help capturing data. Elena Garces was partially supported by a Juan de la Cierva - Incorporacion Fellowship (IJC2020-044192-I).